\documentclass[review]{elsarticle}

\usepackage{color}
\usepackage{multirow} 
\usepackage{amsmath}
\usepackage{amsfonts,amssymb}
\usepackage{float}
\usepackage{url}

\journal{Journal of \LaTeX\ Templates}

\begin{document}

\begin{frontmatter}

\title{TIPCB: A Simple but Effective Part-based Convolutional Baseline for Text-based Person Search}

\author[nuist]{Yuhao Chen}
\ead{chinayhchen@gmail.com}

\author[nuist,massey]{Guoqing Zhang\corref{myfootnote}}
\ead{xiayang14551@163.com}

\author[nuist]{Yujiang Lu}
\ead{yujiang_lu@163.com}

\author[math]{Zhenxing Wang}
\ead{wangzx13142021@163.com}

\author[nuist]{Yuhui Zheng}
\ead{zheng_yuhui@nuist.edu.cn}

\author[massey]{Ruili Wang}
\ead{Ruili.wang@massey.ac.nz}

\address[nuist]{School of Computer and Software, Nanjing University of Information Science and Technology, Nanjing, 210044, China}
\address[math]{School of Mathematics and Statistics, Nanjing University of Information Science and Technology, Nanjing, 210044, China}
\address[massey]{Institute of Natural and Mathematical Sciences, Massey University, Auckland, 4442, New Zealand}
\cortext[myfootnote]{Corresponding author}




\begin{abstract}
Text-based person search is a sub-task in the field of image retrieval, which aims to retrieve target person images according to a given textual description. The significant feature gap between two modalities makes this task very challenging. Many existing methods attempt to utilize local alignment to address this problem in the fine-grained level. However, most relevant methods introduce additional models or complicated training and evaluation strategies, which are hard to use in realistic scenarios. In order to facilitate the practical application, we propose a simple but effective end-to-end learning framework for text-based person search named \textbf{TIPCB} (i.e., \textbf{T}ext-\textbf{I}mage \textbf{P}art-based \textbf{C}onvolutional \textbf{B}aseline). Firstly, a novel dual-path local alignment network structure is proposed to extract visual and textual local representations, in which images are segmented horizontally and texts are aligned adaptively. Then, we propose a multi-stage cross-modal matching strategy, which eliminates the modality gap from three feature levels, including low level, local level and global level. Extensive experiments are conducted on the widely-used benchmark dataset (CUHK-PEDES) and verify that our method outperforms the state-of-the-art methods by \textbf{3.69\%}, \textbf{2.95\%} and \textbf{2.31\%} in terms of Top-1, Top-5 and Top-10. Our code has been released in \textcolor{blue}{\url{https://github.com/OrangeYHChen/TIPCB}}.
\end{abstract}


\end{frontmatter}


\section{Introduction}
Person search is a key technology in the field of image retrieval, aiming to find target person images from large databases with given retrieval conditions, including person images, relevant attributes or natural language descriptions. According to the modality of the query, this technology can be broadly divided into image-based search \cite{1, 2, 3}, attribute-based search \cite{4, 5, 6} and text-based search \cite{7, 8, 9}. In recent years, person search has gained increasing attention due to its wide applications in public security and video surveillance, e.g., searching for suspects and missing persons.

\begin{figure}[!t]
\centering
\includegraphics[width=1\columnwidth]{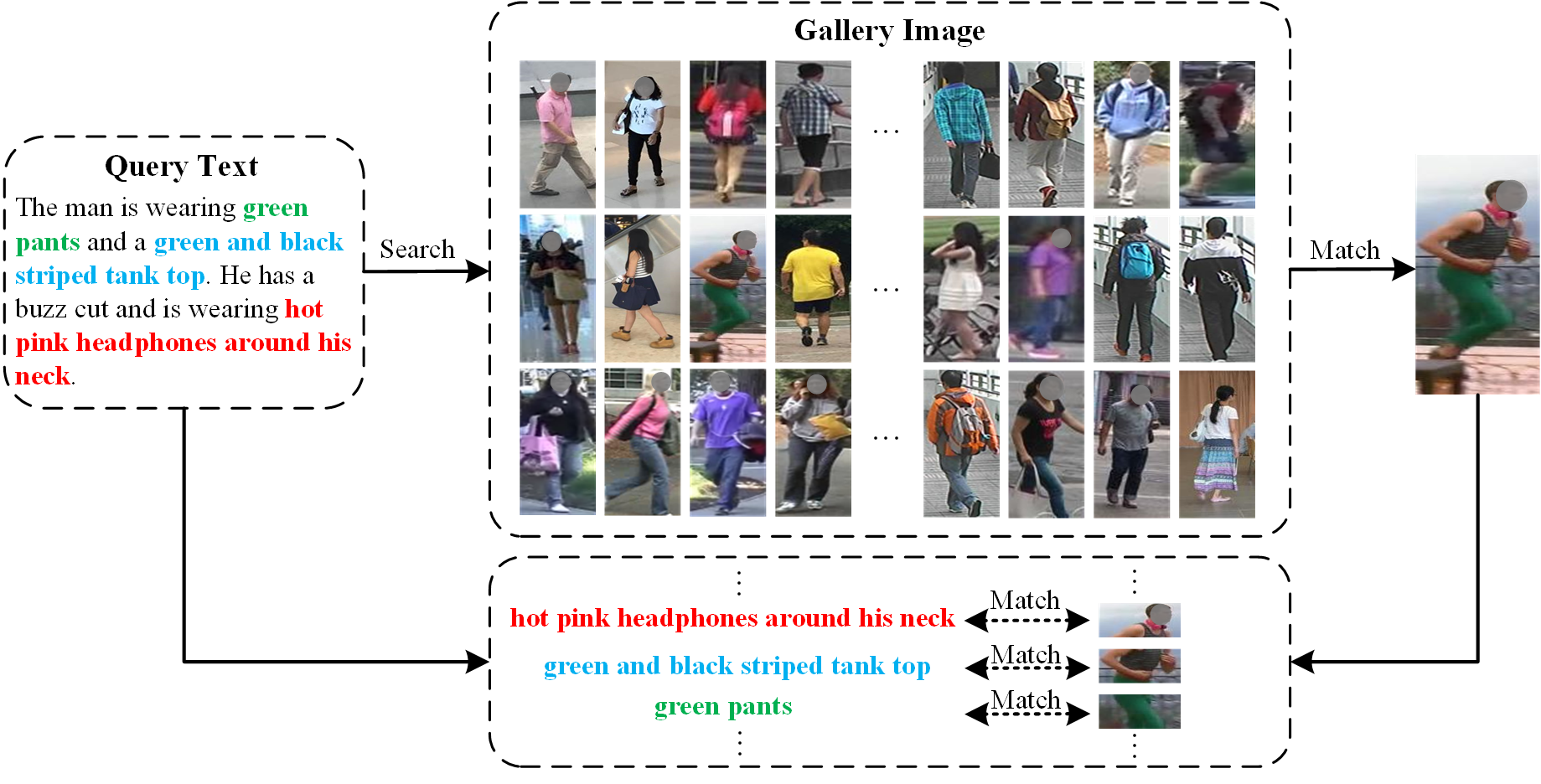}\\
\caption{Illustration of text-based person search problem. Given a textual description, it is aimed to retrieve the corresponding person images from a large gallery database. Since some key information is hidden in the local details, matching local features is necessary for the enhancement of compatibility between person images and textual descriptions.}
\label{fig1}
\end{figure}

In this paper, we research on the task of text-based person search, as is illustrated in Figure \ref{fig1}. Specifically, it is required to sort all person images in a large gallery according to their similarity with the textual description of the query, and select the top person images as matching items \cite{7}. Since textual descriptions are much more natural and accessible as retrieval queries, text-based person search has large potential values in conditions without target person images, e.g., searching for a suspect according to the description of the eyewitness.

Text-based person search is still a challenging task because it has the difficulties of both person re-identification and cross-modal retrieval. On the one hand, it is hard to extract robust visual representations due to the disturbance from occlusion, background clutter and pose/viewpoint variances. On the other hand, some images or descriptions of different persons have very similar high-level semantics, while the domains of image and text have significant differences, resulting in inter-modal feature variances much larger than intra-modal feature variances.

Therefore, a series of relevant methods have been proposed to reduce the gap between image domain and text domain in recent years. We broadly categorize them into global-matching methods and local-matching methods. Global-matching methods mainly focus on the global visual and textual representation learning and obtain the unified feature space regardless of modality \cite{7, 8, 10, 11, 12, 13, 14, 15}. However, images contain many distinctive local details which are hard to explore by global representation extraction. Besides, there are a few irrelevant regions in images, which bring noises to global information. In order to further mine discriminative and comprehensive information, some local-matching methods are proposed, which match person images and textual descriptions by local alignment \cite{9, 16, 17, 18, 19, 20, 21}.

However, most of the existing local-matching methods are not practical enough to meet the requirements of realistic scenarios due to their high complexity. Some of these methods introduce additional models or apply multi-task learning strategies, such as human pose estimation \cite{17, 22, 23}, semantic segmentation \cite{18, 24} or attribute recognition \cite{21, 25}, which bring a huge amount of computation and make networks unable to perform end-to-end learning. Besides, some of these methods adopt the multi-granularities similarity measure strategy \cite{16, 19}. In the use phase, these method need to learn multiple local representations for each image or text, and repeatedly calculate the local similarity. Both additional models and complex similarity measure are time-consuming for practical applications. Thus, it is necessary to design a simple but effective framework for text-based person search problem.

In this paper, we propose a novel end-to-end learning framework named \textbf{TIPCB} (i.e., \textbf{T}ext-\textbf{I}mage \textbf{P}art-based \textbf{C}onvolutional \textbf{B}aseline) to facilitate the practical application. Firstly, a novel dual-path local alignment network structure is proposed to extract visual and textual local representations. Visual local representations are extracted by general PCB strategy \cite{2}, in which person images are horizontally segmented into several stripes. In the textual representation learning path, the word embeddings are learnt by a BERT model with pretrained and fixed parameters \cite{26}, and further processed by a multi-branch residual network. In each branch, the textual representation is learnt to adaptively match a corresponding visual local representation, so as to extract the aligned textual local representation. Besides, a multi-stage cross-modal matching strategy is proposed, which eliminates the modality gap from low-level, local-level and global-level features, and then the feature gap between image domain and text domain can be reduced in a progressive way.

The main contributions of this paper can be summarized as follows:
\begin{itemize}
\item	A novel dual-path local alignment network is proposed to jointly learn visual and textual representations, which can align local features in a simple but effective way.
\item	A multi-stage cross-modal matching strategy is designed to reduce the gap between two modalities in a progressive way. The whole framework can be trained in end-to-end manner.
\item	Extensive experiments are conducted on CUHK-PEDES dataset \cite{7}, and the results clearly verify that our proposed TIPCB framework achieves the state-of-the-art performance.
\end{itemize}

\section{Related Work}
\subsection{Person Re-identification}
In the past decade, a variety of person re-identification (image-based person search) methods have sprung up, attempting to extract discriminative and robust representations for person images, and overcome the difficulties of occlusion, background clutter and so on \cite{27, 28}. Among them, local representation learning is an efficient and widely-used technology, which can explore some detailed and distinctive features. Specifically, PCB \cite{2} applies horizontal segmentation on the feature map of each person image, and extract local representations for obtained stripes independently. MGN \cite{29} adopts a multi-granularities representation learning strategy which cuts each feature map into several parts with different scales. Besides, spatial attention mechanism is introduced to align human parts and further improves the robustness of local representation \cite{30}. Furthermore, some methods utilize additional models to assist local segmentation and mine detailed information, such as pose estimation \cite{31, 32} and human semantic segmentation \cite{33}. 

Extensive works have also achieved great progress in video-based re-ID \cite{34, 35, 36}, occluded re-ID \cite{37}, unsupervised re-ID \cite{38, 39, 40}, cross-resolution re-ID \cite{41, 42, 43}, RGB-IR cross-modality re-ID \cite{44, 45, 46} and so on. Person re-identification has obtained rapid development, but this technology cannot be applied in the scenario without query images.

\subsection{Text-based Person Search}
A series of high-performance methods have been proposed for text-based person search in recent years, which can be broadly categorized into global-matching methods and local-matching methods. GNA-RNN \cite{7} is the first designed framework for this task, which combines a recurrent neural network with the proposed gated neural attention mechanism to learn affinities between textual descriptions and person images. Later, an identity-aware two-stage network \cite{13} is proposed to jointly minimize the intra-identity distance and the cross-modal distance. Besides, in order to embed images and texts to a shared visual-textual space, Zheng et al. \cite{12} design an end-to-end trainable model with a CNN+CNN dual-path structure, and Zhang et al. \cite{14} introduce the cross-modal projection learning in objective functions. Furthermore, TIMAM \cite{15} attempts to learn modality-invariant representations in a shared space by adversarial learning. However, the above methods only focus on global representations, which may miss some distinctive local details or mix a little noise information.

Therefore, some local-matching methods are explored to overcome this shortcoming. Aggarwal et al. \cite{21} introduce human attribute recognition to help bridge the modality gap between the image-text inputs. Wang et al. \cite{18} design a light auxiliary attribute segmentation layer to guide the alignment between visual local representations with parsed textual attributes. Jing et al. \cite{17} propose a multi-granularities attention network to align visual and textual local representations with the aid of the human pose information. These methods apply additional models or multi-task learning strategies to enhance the local alignment but bring a huge amount of computation. In addition, MIA \cite{19} aligns the local representations from multiple granularities, including global-global, global-local and local-local levels. NAFS \cite{16} conducts joint alignments over full-scale representations with a novel staircase CNN and a locality-constrained BERT. Nevertheless, such methods are still time-consuming in the use phase due to their complex similarity measure strategies. By comparison, our proposed method utilizes an end-to-end trainable dual-path network to learn local aligned representations simply and effectively.

\section{Proposed Method}
In this section, we will detailly explain our proposed Text-Image Part-based Convolutional Baseline (TIPCB) for text-based person search problem. We first illustrate the dual-path local alignment network structure, including the visual CNN branch and the textual CNN branch, and then the multi-stage cross-modal matching strategy is introduced to eliminate the modality gap.

\begin{figure}[!t]
\centering
\includegraphics[width=1\columnwidth]{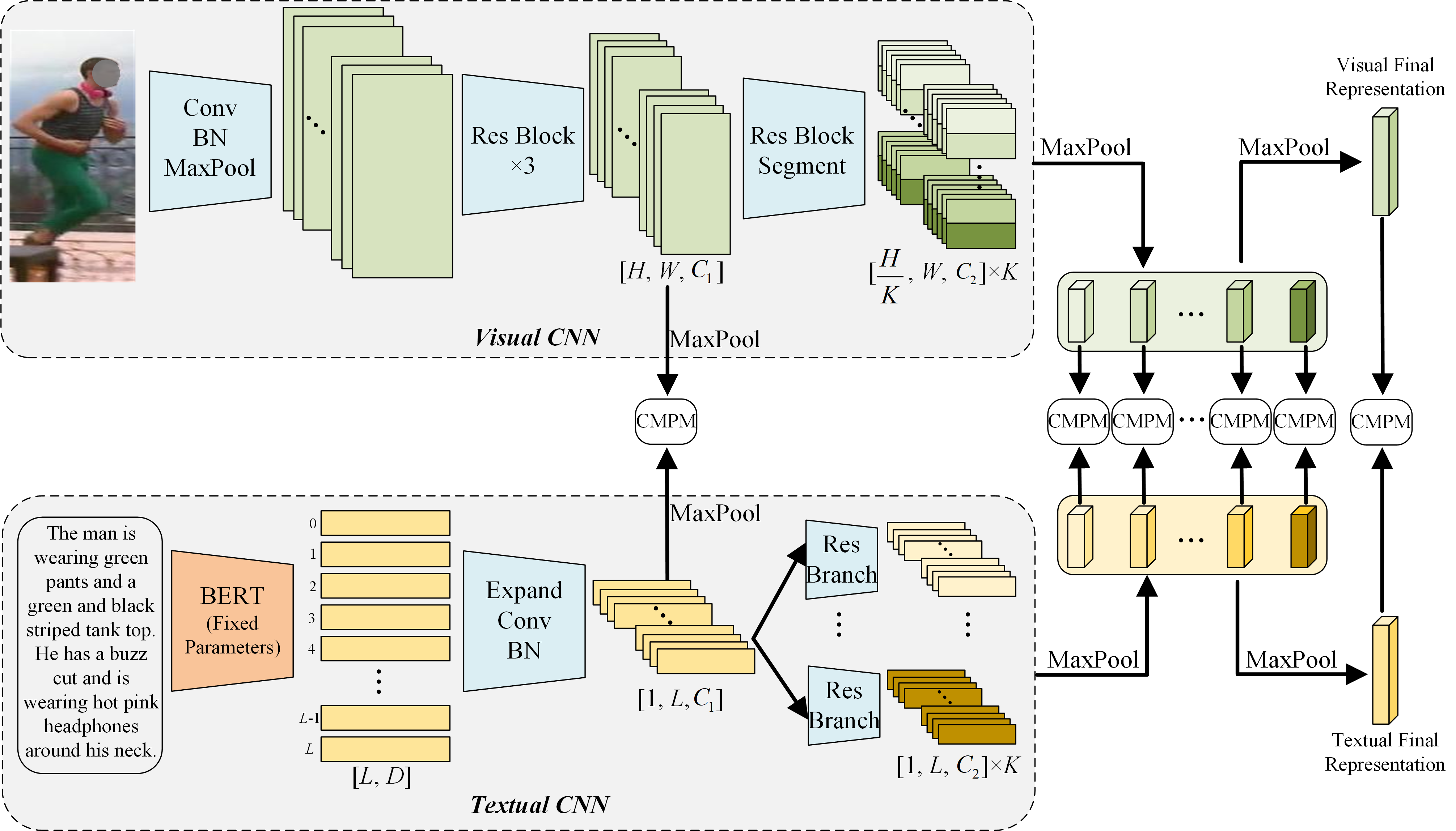}\\
\caption{The architecture of the proposed TIPCB. This framework consists of a dual-path local alignment structure, where the visual CNN branch applies the PCB after the backbone network, and the textual CNN branch applies a multi-branch residual network after a pretrained BERT model. This framework adopts a multi-stage cross-modal matching strategy, which conducts projection matching on low-level, local-level and global-level representations.}
\label{fig2}
\end{figure}

\subsection{Visual Representation Learning}
As illustrated in Figure \ref{fig2}, our proposed TIPCB contains two CNN branches which aim to learn discriminative and compatible visual and textual representations from the input person images and descriptions, respectively. In the training phase, we assume the training data as $D{\rm{ = }}\left\{ {{I_i},{T_i}} \right\}_{i = 1}^N$ where $N$ represents the number of image-text pair in each batch, and each pair consists of an image $I$ and a corresponding description $T$. (The subscript $i$ is omitted in the following for simplicity unless necessary.)
In visual CNN branch, ResNet-50 \cite{47} is adopted as the backbone to extract visual features, which mainly consists of four residual blocks. Different residual blocks can capture semantic information from different level \cite{48}. For each image $I$, we define the feature generated by the ${3^{{\rm{rd}}}}$ and ${4^{{\rm{th}}}}$ residual block as its low-level feature map $f_l^I \in {\mathbb{R}^{H \times W \times {C_1}}}$ and high-level feature map $f_h^I \in {\mathbb{R}^{H \times W \times {C_2}}}$, where $H$, $W$ and ${C_1}$/${C_2}$ represent the dimension of height, width and channel in the above feature maps. Then we obtain its visual low-level representation $v_l^I \in {\mathbb{R}^{{C_1}}}$ by:
\begin{equation}
    v_l^I = {\rm{GMP}}(f_l^I)
\label{f1}
\end{equation}
where GMP represents a global max-pooling layer as a filter to mine salient information. 

Here, we adopt the PCB \cite{2} strategy to obtain the local regions. Specially, the high-level feature map $f_h^I$ is segmented into $K$ horizontal stripes which are denoted as $\left\{ {f_{p1}^I,f_{p2}^I,...,f_{pK}^I} \right\}$ where $f_{pi}^I \in {\mathbb{R}^{\frac{H}{K} \times W \times {C_2}}}$. For each stripe, similar to Formula (\ref{f1}), we still adopt a global max-pooling layer to extract the visual local representation $v_{pi}^I \in {\mathbb{R}^{{C_2}}}$. In order to fuse all local representations, we select the maximum value of each element in channel dimension, and get the visual global representation $v_g^I \in {\mathbb{R}^{{C_2}}}$:
\begin{equation}
    v_g^I = {\rm{Max}}\left( {v_{p1}^I,v_{p2}^I,...,v_{pK}^I} \right)
\label{f2}
\end{equation}
Therefore, we get the visual feature set ${V^I} = \left\{ {v_l^I,v_{p1}^I,...,v_{pK}^I,v_g^I} \right\}$ containing low-level, local-level and global-level representations. In the testing phase, only the global-level representation is adopted to measure similarity.

\subsection{Textual Representation Learning}
In textual CNN branch, a high-performance language representation model BERT \cite{26} is applied to extract discriminative word embeddings, which can learn the contextual relations between the words by the bi-directional training of Transformer \cite{49}. Specifically, we break each textual description $T$ up into a list of words, and insert [CLS] and [SEP] into the beginning and end of each sentence. Then this list is embedded into tokens by a pretrained tokenizer. To ensure the consistency of text length, we select the first $L$ tokens when the text is longer than $L$, and apply zero-padding in the end of the text when the text is shorter than $L$. After that, each tokenized textual description is input into the BERT model, which is pretrained and parameter-fixed, to extract word embeddings $t \in {\mathbb{R}^{L \times D}}$, where $D$ represents the dimension of each word embedding. Here we “freeze” the weights of the BERT model for the following three reasons: 1) the pretrained BERT itself has the strong semantic representation ability, and we only use it as the word embedding layer, 2) the following CNN structure is capable to further process the word embeddings, and 3) only training the CNN structure can significantly reduce the amount of training parameters and accelerate the convergence of the model.

\begin{figure}[!t]
\centering
\includegraphics[width=0.9\columnwidth]{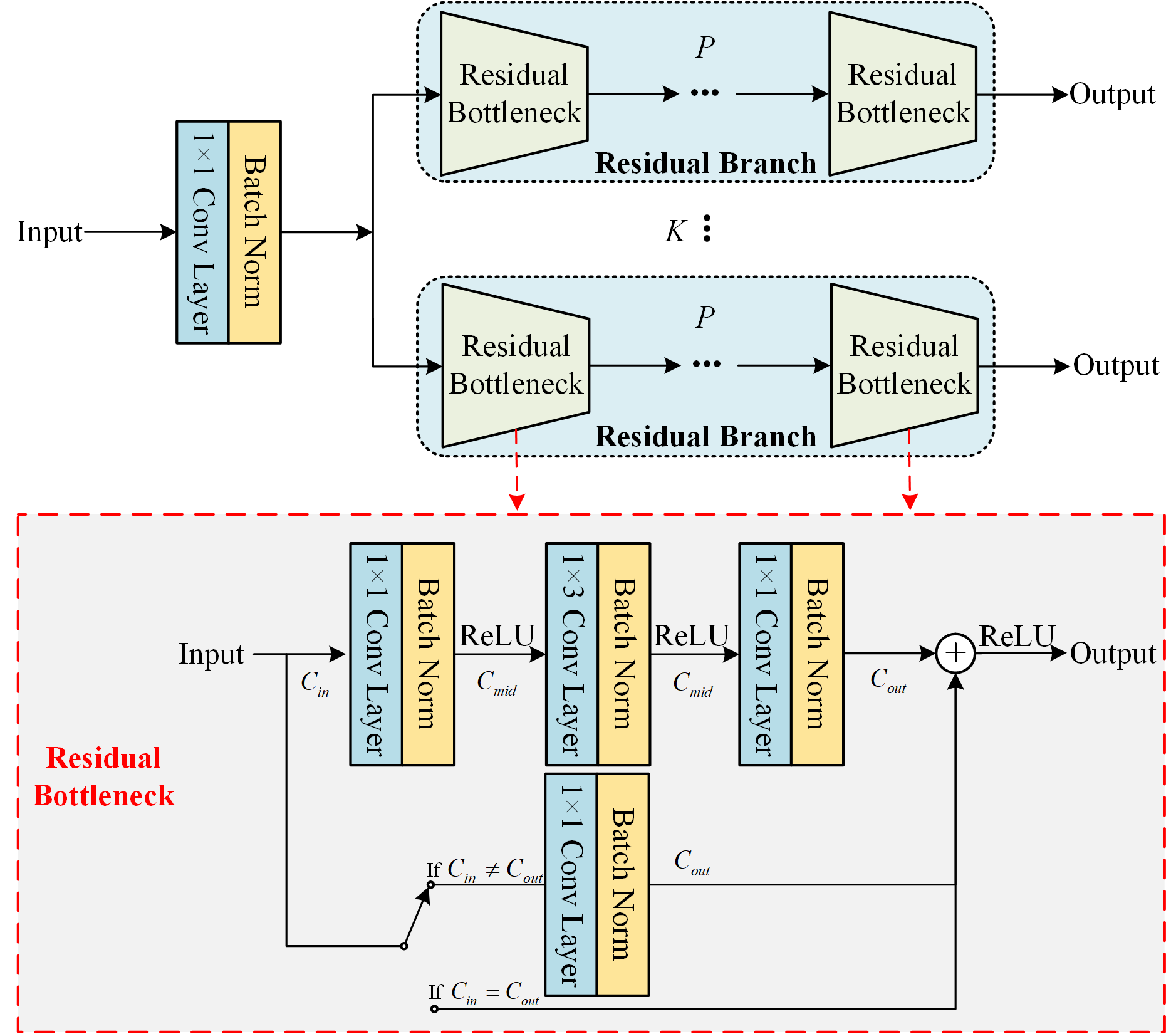}\\
\caption{The details of the multi-branch textual CNN.}
\label{fig3}
\end{figure}

In order to meet the inputting requirement of convolutional layers, we expand the dimension of word embedding from $t \in {\mathbb{R}^{L \times D}}$ to ${t^ * } \in {\mathbb{R}^{1 \times L \times D}}$, where 1, $L$ and $D$ are regarded as the height, width and channel dimension of convolutional input, respectively. Motivated by the residual network \cite{47} and the deep textual CNN \cite{12}, we design the multi-branch textual CNN, as shown in Figure \ref{fig3}. In the textual CNN, in order to map the word embeddings to the same channel dimension as the visual low-level feature map $f_l^I \in {\mathbb{R}^{H \times W \times {C_1}}}$, the filter size of the first convolutional layer is set to $1 \times 1 \times D \times {C_1}$, which can be viewed as a lookup table. Then, we can obtain the textual low-level feature map $f_l^T \in {\mathbb{R}^{1 \times L \times {C_1}}}$. 

The multi-branch textual CNN contains $K$ residual branches, which correspond to the $K$ stripes of person images. For each branch, it contains $P$ textual residual bottlenecks and is aimed to adaptively learn the textual representations which can match the visual local representations. The textual residual bottleneck has the similar structure as the modules in ResNet, consisting of several convolutional layers and batch normalization layers. The skip connection is applied to transmit information from low layers to high layers, which can effectively restrain the network degradation problem and speed up the model training. Specifically, in order to keep the textual information uncompressed,
the strides of all convolutional layers in bottlenecks are set to $1 \times 1$. 
For the first bottleneck of each branch, we modify the channel dimension of the textual feature map to ${C_2}$, which is consistent with the visual high-level feature map $f_h^I \in {\mathbb{R}^{H \times W \times {C_2}}}$, and then we keep the channel dimension unchanged in the following bottlenecks. After the multi-branch textual CNN, we obtain the textual local feature maps. Similar to visual CNN branch, we adopt a global max-pooling layer to extract the textual local representations and select the maximum value of each element in channel dimension to fuse these local representations. Then, we get the textual feature set ${V^T} = \left\{ {v_l^T,v_{p1}^T,...,v_{pK}^T,v_g^T} \right\}$ containing low-level, local-level and global-level representations.

Different from the deep textual CNN \cite{12}, we only stack a few bottlenecks rather than use a very deep residual network to extract textual representations for the following two reasons: 1) the downsampling between different stages in deep textual CNN brings obvious information loss, and 2) deep network does not bring obvious improvement compared with shallow network, which is contrary to the experience in image area and has been proven in \cite{50}. In our experiment section, we will further verify the above viewpoints.

\subsection{Multi-stage Cross-modal Matching}
In order to eliminate the feature gap between image modality and text modality, we adopt the Cross-Modal Projection Matching (CMPM) loss \cite{14} on low-level, local-level and global-level representations, which can associate representations across different modalities by incorporating the cross-modal projection into KL divergence. For each visual representation $v_i^I$, we assume that the set of image-text representation pairs is $\left\{ {\left( {v_i^I,v_j^T} \right),{y_{i,j}}} \right\}_{j = 1}^N$, where ${y_{i,j}} = 1$ represents that $v_i^I$ and $v_j^T$ are from the same person, while ${y_{i,j}} = 0$ means that they are not a matched pair. (The subscript used to indicate the representation level is omitted, because it is applicable for any representation pair from ${V^I}$ and ${V^T}$.)

The probability that $v_i^I$ and $v_j^T$ are a matched pair can be calculated by:
\begin{equation}
   {p_{i,j}} = \frac{{\exp \left( {{{(v_i^I)}^ \top }\bar v_j^T} \right)}}{{\sum\limits_{k = 1}^N {\exp \left( {{{(v_i^I)}^ \top }\bar v_k^T} \right)} }}
\label{f3}
\end{equation}
where $\bar v_j^T$ is the normalized textual representation and denoted as $\bar v_j^T = \frac{{v_j^T}}{{\left\| {v_j^T} \right\|}}$. In CMPM, the scalar projection of $v_i^I$ on $v_j^T$ is regarded as their similarity, and matching probability ${p_{i,j}}$ is the proportion of the similarity between $v_i^I$ and $v_j^T$ to the sum of similarity between $v_i^I$ and $\left\{ {v_j^T} \right\}_{j = 1}^N$ in a batch. Then the CMPM loss can be calculated by:
\begin{equation}
   {L_{I2T}} = \frac{1}{N}\sum\limits_{i = 1}^N {\sum\limits_{j = 1}^N {{p_{i,j}}\log \left( {\frac{{{p_{i,j}}}}{{{q_{i,j}} + \varepsilon }}} \right)} }
\label{f4}
\end{equation}
where $\varepsilon$ is a small number to avoid numerical problems, and ${q_{i,j}}$ is the normalized true matching probability between $v_i^I$ and $v_j^T$ since there might be more than one matched text descriptions in a batch, denoted as ${q_{i,j}} = \frac{{{y_{i,j}}}}{{\sum\limits_{k = 1}^N {{y_{i,k}}} }}$. The above procedure reduces the distance between each visual representation and its matched textual representations in a single direction, and we reversely conduct the similar procedure to draw each textual representation and its matched visual representations closer. Therefore, the bi-directional CMPM loss is computed by:
\begin{equation}
   {L_{CMPM}} = {L_{I2T}} + {l_{T2I}}
\label{f5}
\end{equation}

The objective in our framework contains the cross-modal representation matching from three levels. The CMPM loss in low-level representations is to reduce the modality gap in an early stage. The CMPM loss in local-level representations can realize the local alignment between images and texts. The CMPM loss in global-level representations ensure that the final representations for evaluation have the stronger modal compatibility. Through the multiple stages of CMPM loss, the matching degree of image-text representations can be gradually improved, which will be further verified in ablation study. Finally, according to the visual and textual representation sets ${V^I}$ and ${V^T}$, the overall objective function is calculated by:
\begin{equation}
   L = {\lambda _1}L_{CMPM}^l + {\lambda _2}\sum\limits_{k = 1}^K {L_{CMPM}^{pk}}  + {\lambda _3}L_{CMPM}^g
\label{f6}
\end{equation}
where ${\lambda _1}$, ${\lambda _2}$, ${\lambda _3}$ are hyper-parameters to control the importance of different CMPM losses, and $L_{CMPM}^l$, $\left\{ {L_{CMPM}^{pk}} \right\}_{k = 1}^K$, $L_{CMPM}^g$ indicate the CMPM loss of low-level, local-level and global-level representations, respectively.

\section{Experiment}
\subsection{Experimental Settings}
\begin{figure}[!t]
\centering
\includegraphics[width=1\columnwidth]{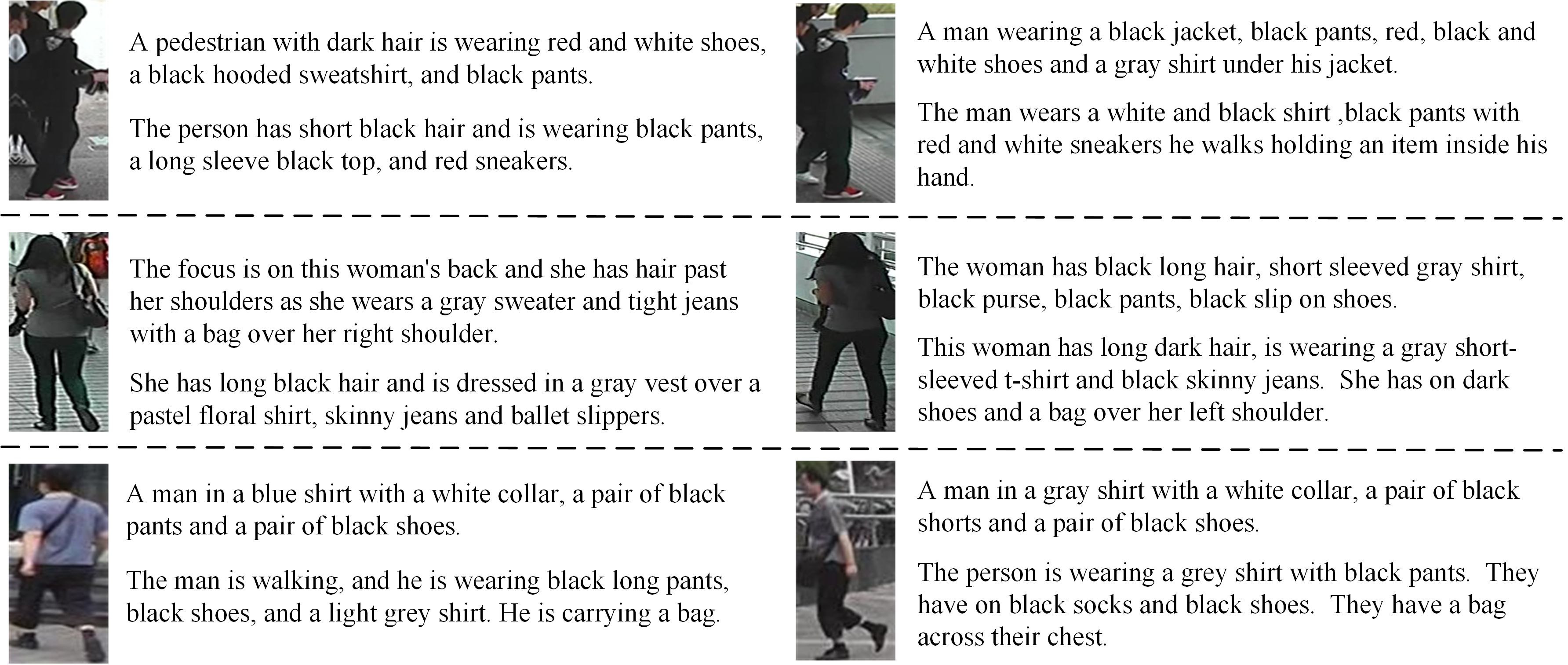}\\
\caption{Samples in the CUHK-PEDES dataset. The person images in each line are of the same identity, and each image has two corresponding textual descriptions.}
\label{fig4}
\end{figure}

\textbf{Dataset.} We evaluate our proposed TIPCB framework on the CUHK-PEDES dataset \cite{7}, which is the only large scale available benchmark for text-based person search problem, as is shown in Figure \ref{fig4}. It totally contains 40,206 person images of 13,003 identities by collecting samples from several re-ID datasets. Each person image has two corresponding textual descriptions on average, and each textual description has more than 23 words. These textual descriptions have a vocabulary of 9,408 different words. We adopt the same data split as \cite{7}. The training set has 34,054 images of 11,003 identities. The validation and testing set have 3,078 images and 3,074 images of 1,000 identities, respectively.

\textbf{Evaluation Protocol.} We follow the standard evaluation metrics, and report the top-$k$ ($k = 1,5,10$) accuracy to evaluate the performance. Specifically, given a query text description, all gallery images are ranked according to their similarity values. A successful search means that a matched person image is existed among the top-$k$ images.

\textbf{Implementation Details.} In the visual CNN branch, we adopt ResNet-50 pretrained on ImageNet \cite{51} as the backbone to extract visual feature maps, and we modify the stride of $conv5\_1$ to 1 instead of 2 for larger feature maps. In the textual CNN branch, we extract word embeddings by the language model BERT-Base-Uncase pretrained on a large corpus including Toronto Book Corpus and Wikipedia. All the input images are resized to $384 \times 128$, and the text length is unified to $L = 64$. The number of local regions is set to $K{\rm{ = }}6$. In the multi-branch textual CNN, the number of bottlenecks in each residual branch is set to $P{\rm{ = 3}}$. In the visual and textual features, some parameters of dimensions are set to $H = 24$, $W = 8$, $D = 768$, ${C_1} = 1024$ and ${C_2} = 2048$. Each batch contains $N = 64$ image-text pairs. 

In the training phase, Adam is selected to optimize our model with weight decay $4 \times {10^{ - 5}}$. The model is trained for 80 epochs in total. The base learning rate is set to $3 \times {10^{ - 3}}$ and decreased by 0.1 after 50 epochs. Besides, we initialize the learning rate by the warm-up trick in first 10 epochs. We adopt the trick of horizontally flipping to augment data, where each image has 50\% chance to flip randomly. The hyper-parameters in the objective function are set to ${\lambda _1} = 1$, ${\lambda _2} = 1$ and ${\lambda _3} = 1$. In the testing phase, the cosine distance is used to measure the similarity value of image-text pairs. We perform our experiments with PyTorch on a single Tesla V100 GPU.

\subsection{Comparison with State-of-The-Art Methods}
We compare our method with the existing text-based person search methods, and the comparable results are reported in Table \ref{tab1}. We divide these methods into two categories. Global-matching methods (type column is marked as “G”) consist of GNA-RNN, IATV, Dual Path, CMPM + CMPC and TIMAM, and local-matching methods (type column is marked as “L”) contain PWM + ATH, GLA, MIA, PMA, ViTAA, CMAAM and NAFS. We can find that the methods based on local alignment have become a hot topic and relatively achieved better performance in recent years, which can prove the importance of the local fine-grained alignment to strengthen the compatibility and discrimination of image-text features.

\begin{table}[!t]
\caption{Comparison with state-of-the-art methods on CUHK-PEDES dataset. Top-1, Top-5 and Top-10 accuracies (\%) are reported. The ${1^{{\rm{st}}}}$, ${2^{{\rm{nd}}}}$ and ${3^{{\rm{rd}}}}$ top results are indicated by \textcolor{red}{\textbf{red}}, \textcolor{blue}{\textbf{blue}} and \textcolor{green}{\textbf{green}} bold numbers, respectively. In the second column, “G” represents the methods only using global features, and “L” indicates the methods aligning local features. All listed methods do not use post-processing including re-ranking, to ensure the fairness of comparison.} 
\centering
\label{tab1}
\resizebox{0.9\textwidth}{!}{
\begin{tabular}{c c c c c c}
\hline 
Method & Type & Ref & Top-1 & Top-5 & Top-10 \\
\hline
GNA-RNN \cite{7} & G & CVPR17 & 19.05 & - & 53.64 \\
IATV \cite{17} & G & ICCV17 & 25.94 & - & 60.48 \\
PWM+ATH \cite{19} & L & WACV18 & 27.14 & 49.45 & 61.02 \\
GLA \cite{21} & L & ECCV18 & 43.58 & 66.93 & 76.26 \\
Dual Path \cite{15} & G & TOMM20 & 44.40 & 66.26 & 75.07 \\
CMPM+CMPC \cite{19} & G & ECCV18 & 49.37 & - & 79.27 \\
MCCL \cite{10} & G & ICASSP19 & 50.58 & - & 79.06 \\
MIA \cite{16} & L & TIP20 & 53.10 & 75.00 & 82.90 \\
PMA \cite{12} & L & AAAI20 & 53.81 & 73.54 & 81.23 \\
ViTAA \cite{13} & L & ECCV20 & 55.97 & 75.84 & 83.52 \\
CMAAM \cite{22} & L & WACV20 & \textcolor{green}{\textbf{56.68}} & 77.18 & \textcolor{green}{\textbf{84.86}} \\
\hline
\multicolumn{6}{c}{ResNet+BERT Structure}  \\
\hline
TIMAM \cite{20} & G & ICCV19 & 54.51 & \textcolor{green}{\textbf{77.56}} & 84.78 \\
NAFS \cite{11} & L & arXiv21 & \textcolor{blue}{\textbf{59.94}} & \textcolor{blue}{\textbf{79.86}} & \textcolor{blue}{\textbf{86.70}} \\
TIPCB (ours) & L & PR21 & \textcolor{red}{\textbf{63.63(↑3.69)}} & \textcolor{red}{\textbf{82.82(↑2.95)}} & \textcolor{red}{\textbf{89.01(↑2.31)}} \\
\hline
\end{tabular}}
\end{table}

\subsection{Ablation Studies}
\begin{figure}[!t]
\centering
\includegraphics[width=1\columnwidth]{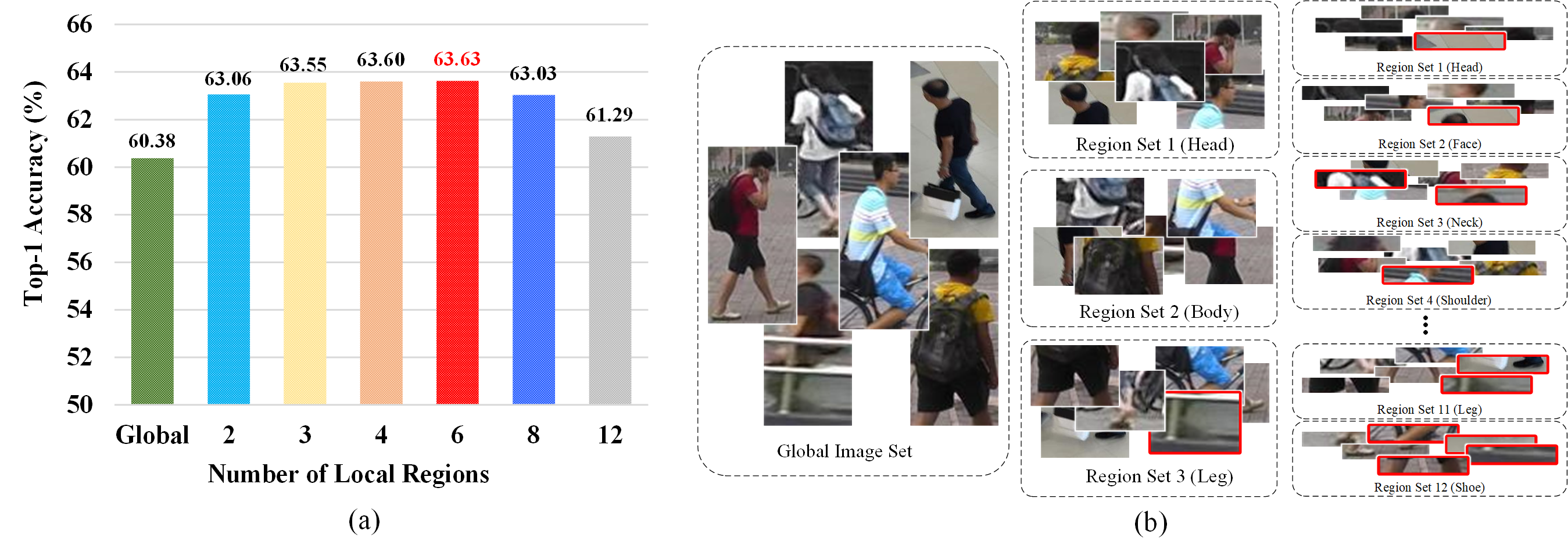}\\
\caption{(a) Comparable results of different number of regions. Top-1 accuracy (\%) is reported. (b) The region sets of different scales. The noises inside are marked by red bounding boxes.}
\label{fig5}
\end{figure}

\textbf{Effects of local representations.} In order to verify the effectiveness of local representations, we conduct ablation studies to compare the performance of global representations and local representations of different region scales, and the results are listed in Figure \ref{fig5}(a). We can obtain the following two findings. On the one hand, compared with the global representations, the local representations have the much stronger discriminative ability, and the method with 6 local regions gains 3.25\% improvement in Top-1 accuracy. One important reason is that global representations are hard to capture some distinctive local details. On the other hand, it can be observed that the Top-1 accuracy increases first and then decreases with the increase of the number of local regions, which can be explained by the local misalignment among person images. Due to the unstable vibration of person detection boxes and variances of viewpoints, the spatial distribution of head, body and limbs exists significant differences. We apply the hard segmentation strategy based on PCB \cite{2} in the visual CNN branch, which brings the noises in local region set. When the granularity of the local regions is too small, abundant noises in local region set will bring difficulties for the network to extract the common features of this region, as is shown in Figure \ref{fig5}(b).

\begin{figure}[!t]
\centering
\includegraphics[width=1\columnwidth]{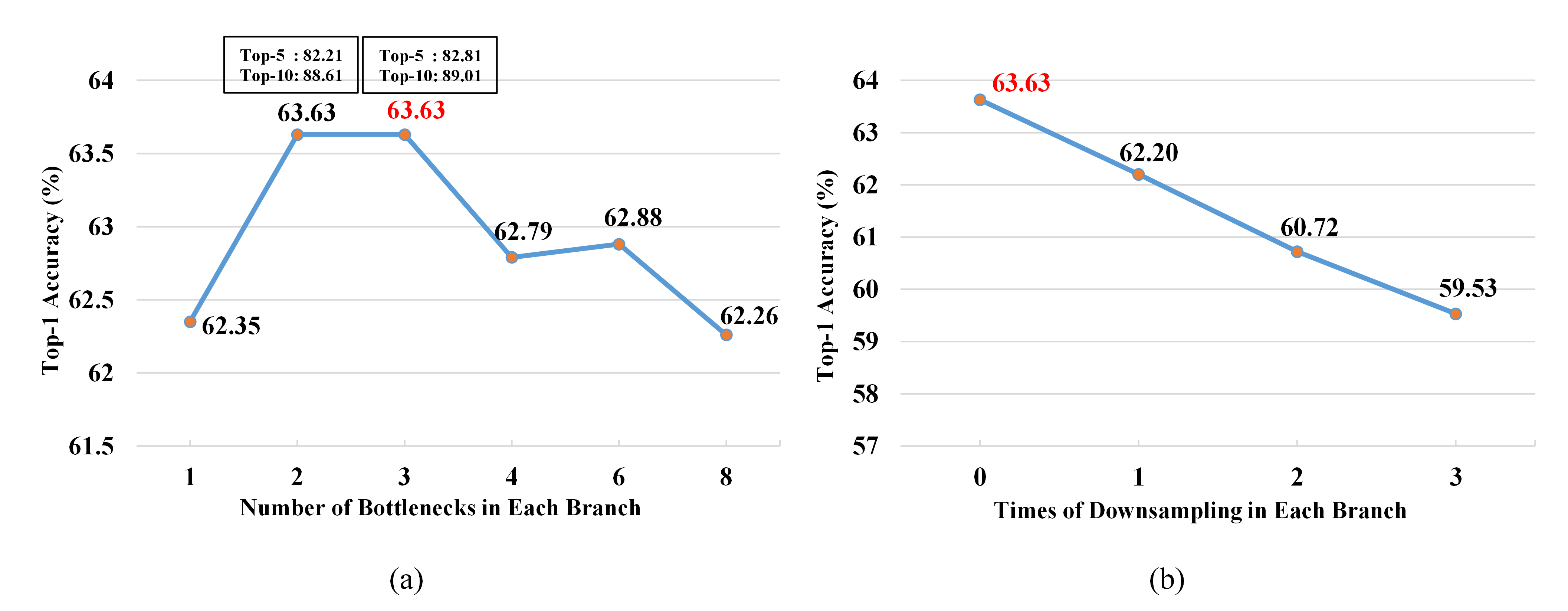}\\
\caption{Comparable results of different number of bottlenecks (in Figure 6(a)) and different times of downsampling in each branch (in Figure 6(b)). Top-1 accuracy (\%) is reported.}
\label{fig6}
\end{figure}

\textbf{Effects of shallow network for textual representation learning.} Motivated by the conclusion that deep network does not bring obvious improvement for text classification compared with shallow network in \cite{50}, we design a much shallower network than the deep textual CNN \cite{12} to extract textual representations from word embeddings. We conduct a series of ablation experiments to prove the superiority of shallow network in textual representation learning. As is shown in Figure \ref{fig6}(a), we test the multi-branch textual CNNs with different number of bottlenecks in each branch. The result presents an overall trend of first increasing and then decreasing with the increase of the number of bottlenecks, and the network achieves the best performance when each branch has 3 bottlenecks. When it has only one residual bottleneck, the network has insufficient ability to extract discriminative features. Besides, the network for textual representation learning does not need too deep layers since text data is generally discrete and sparse, which is significantly different from image data. Therefore, our multi-branch textual CNN performs better in the condition that each branch has only 2$\sim$3 bottlenecks.

In addition, we keep the size of textual feature maps rather than use multi-stage downsampling in \cite{12}. We compare the residual branches with different times of downsampling and the results are shown in Figure \ref{fig6}(b). Specifically, each residual branch has 3 residual bottlenecks by default, and the branch without downsampling outputs the feature map of $1 \times 64 \times 2048$. The spatial size of feature map will be reduced by half when each bottleneck applies the downsampling strategy. That is to say, the branch with 1, 2 and 3 times of downsampling output the feature maps of $1 \times 32 \times 2048$, $1 \times 16 \times 2048$ and $1 \times 8 \times 2048$, respectively. It can be observed that the Top-1 accuracy decreases significantly with the increase of times of downsampling. One of the important reasons is that the downsampling strategy brings obvious textual information loss. Therefore, we do not adopt downsampling and keep the size of textual feature maps unchanged in residual branches.

\textbf{Effects of multi-stage cross-modal matching strategy.} During the training phase, in order to stimulate the modality gap step by step, we apply a multi-stage cross-modal matching strategy, which applies the CMPM loss on the representations of three stages, including low-level and high-level representations. Note that both local-level and global-level representations belong to high-level representations. We conduct the following ablation experiments to verify the CMPM loss in each stage, and the results are reported in Table \ref{tab2}. (The accuracy of the method with only low-level CMPM loss is not listed here, because we only select the high-level representations during the testing phase, and the result will be not referential if the high-level CMPM loss is ignored.)

\begin{table}[!t]
\caption{Comparable results of combinations of CMPM loss in different stages. Top-1, Top-5 and Top-10 accuracies (\%) are reported.} 
\centering
\label{tab2}
\resizebox{0.8\textwidth}{!}{
\begin{tabular}{c c c c c c c}
\hline 
Variant & Low-level & Local-level & Global-level & Top-1 & Top-5 & Top-10 \\
\hline
(a) &  & \checkmark &  & 57.16 & 78.03 & 84.94 \\
(b) &  &  & \checkmark & 58.47 & 80.19 & 86.97 \\
(c) & \checkmark & \checkmark &  & 58.07 & 79.12 & 85.70 \\
(d) & \checkmark &  & \checkmark & 59.84 & 81.39 & 87.95 \\
(e) &  & \checkmark & \checkmark & 62.39 & 81.93 & 88.71 \\
(f) & \checkmark & \checkmark & \checkmark & \textbf{63.63} & \textbf{82.81} & \textbf{89.01} \\
\hline
\end{tabular}}
\end{table}

From the experimental results, we can observe the following findings: 

1)	According to the results of variants (a) and (b) which only use one stage of CMPM loss, we find that their Top-1 accuracies of more than 57\% have already surpasses most of the existing methods except for NAFS. It proves the high efficiency of our dual-path local alignment network.

2)	According to the results of variants (a), (b), (c) and (d), we observe that the addition of low-level CMPM loss brings 0.91\% and 1.37\% improvements in Top-1 accuracy for variants (a) and (b), respectively. It proves the positive effect of matching low-level representations, which can reduce the modality gap in an early stage.

3)	From the variants (a), (b) and (e), it can be found that the combination of local-level and global-level CMPM losses has 5.23\% and 3.92\% improvements in Top-1 accuracy for single local-level and single global-level CMPM loss. The reason is that the local-level CMPM loss can realize the local alignment between images and texts, and the global-level CMPM loss can ensure that the final representations have the stronger modal compatibility. This group of experiments can simultaneously verify the validity of matching local-level and global-level representations.

4)	From the variant (f), we can find that the combination of low-level, local-level and global-level CMPM losses achieves the best performance, which further shows the high efficiency of our proposed multi-stage cross-modal matching strategy.

\textbf{Effects of fusion strategy.} We conduct a series of ablation experiments to compare the performance of different fusion strategies, including avg-pooling, max-pooling and the addition of them, and the results are listed in Table \ref{tab3}. It can be observed that the accuracy of the strategy with max-pooling is far higher than the strategy with avg-pooling, and sightly higher than the strategy with the addition of them. Here, global max-pooling layer can filter salient information to extract more discriminative representation, while global avg-pooling may mix irrelevant noises in representations since it integrates each element in the feature map.

\begin{table}[!t]
\caption{Comparable results of different fusion strategies. Top-1, Top-5 and Top-10 accuracies (\%) are reported.} 
\centering
\label{tab3}
\resizebox{0.6\textwidth}{!}{
\begin{tabular}{c c c c}
\hline 
Fusion Strategy & Top-1 & Top-5 & Top-10 \\
\hline
Avg-pooling & 55.63 & 76.59 & 84.14 \\
Max-pooling & \textbf{63.63} & \textbf{82.81} & \textbf{89.01} \\
Max-pooling+Avg-pooling & 63.22 & 82.73 & 88.21 \\
\hline
\end{tabular}}
\end{table}

\begin{figure}[!t]
\centering
\includegraphics[width=1\columnwidth]{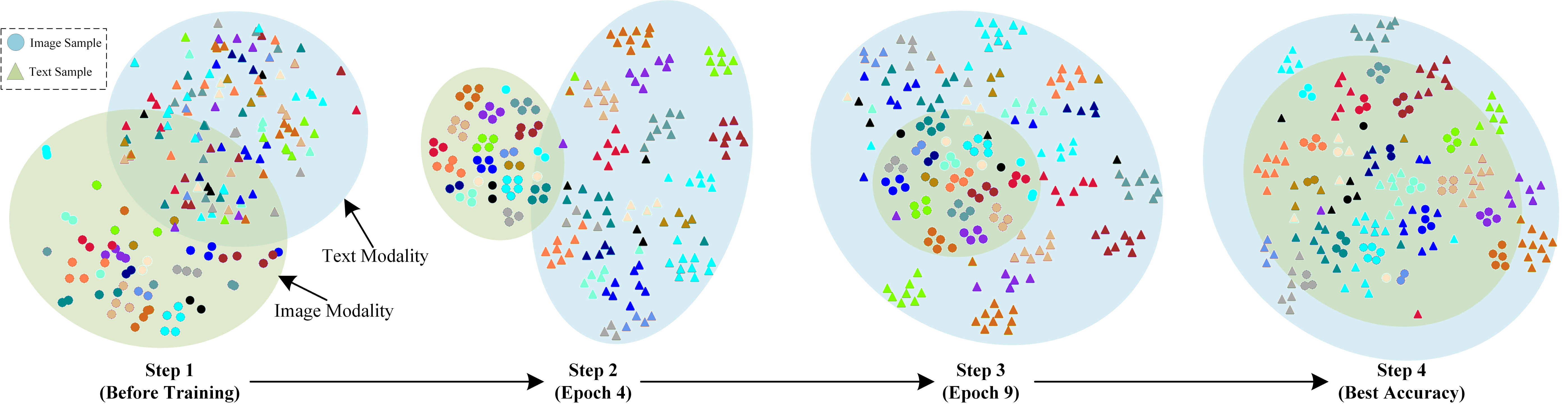}\\
\caption{Visualization of features via t-SNE. We show the changing process of cross-modal feature distributions with training. There are shown 64 images with corresponding 128 textual descriptions. The feature of each image and text is marked as a circle and a rectangle, respectively. Each identity is indicated in a specific color.}
\label{fig7}
\end{figure}

\textbf{Visualizaton of features.} We apply the t-SNE \cite{52} to visualize the features in Figure \ref{fig7} and show the changing process of feature distributions in four steps. Before training, there is a significant gap between text modality and image modality, and the distributions inside the modalities are disordered. After several training epochs, it can be observed that samples of the same identity begin to cluster, but the two modalities still have a large gap. Then, the distributions of two modalities begin to converge gradually until their centers are close. Finally, the feature distributions of two modalities coincide well to some extent, and the samples from the same identity can have a good clustering performance. This demonstrates that our TIPCB is capable to learn discriminative visual and textual representations, and eliminate the cross-modal distribution gap.

\begin{figure}[!t]
\centering
\includegraphics[width=1\columnwidth]{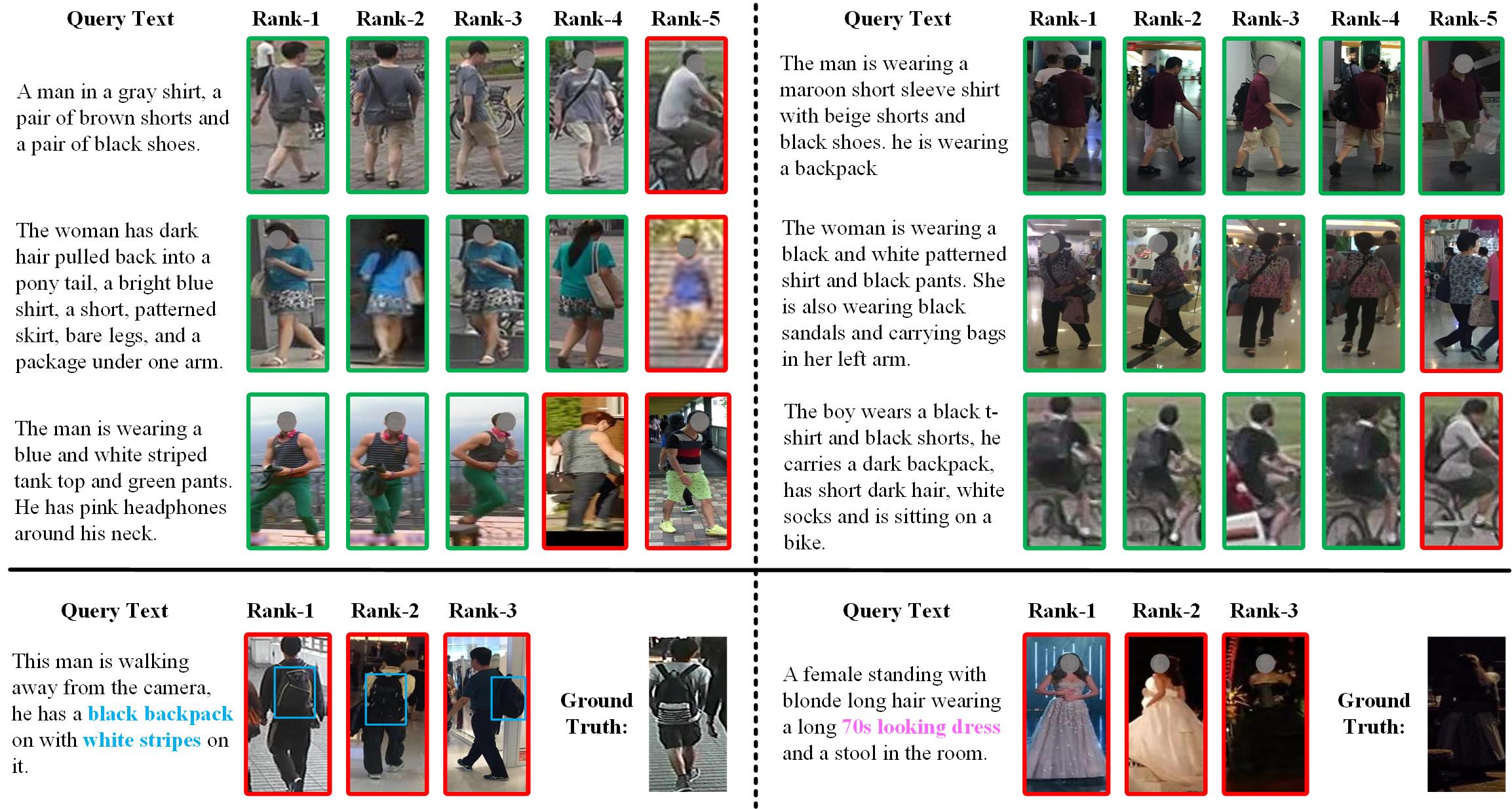}\\
\caption{Visualization of text-based person search results by our proposed TIPCB. The green and red bounding boxes indicate correct and incorrect matches, respectively. The top 6 groups are successful searches, while the bottom 2 groups are failure searches.}
\label{fig8}
\end{figure}

\textbf{Sample analysis.} As is shown in Figure \ref{fig8}, we visualize and analyze several examples of text-based person search by our proposed TIPCB. For each successful case, we can find that the listed top-5 person images have multiple regions that can well match parts of the corresponding textual description. Note that our proposed TIPCB is capable to distinguish some hard samples, which only partly match the textual description. For example, in the case 6, our method successfully finds the perfect matched images which simultaneously meet the characteristics of “black short”, “dark backpack”, “short dark hair” and “sitting on a bike”, and lists the image that only mismatches the condition of “black short” behind.

In terms of failure cases, we list the following two representative examples. In the first case, the textual description is too ambiguous, which only contains useful information about backpack. We can find that the top-3 person images all have the “black backpack”, but they are not the matched ones. Besides, the detailed description of “white stripes” is too subtle to extract discriminative features from it. In the second case, the text uses “70s looking” to describe the dress, which is a rare phase and is difficult for network to learn. Thus, the top-3 person images have the dresses with different styles.

\section{Conclusion}
In this paper, in order to facilitate the practical application, we propose a simple but effective end-to-end learning framework for text-based person search named \textbf{TIPCB} (i.e., \textbf{T}ext-\textbf{I}mage \textbf{P}art-based \textbf{C}onvolutional \textbf{B}aseline). In contrast to the existing local-matching methods, TIPCB applies an end-to-end trainable structure without additional models and complex evaluation strategies. We design a novel dual-path local alignment network to learn visual and textual local representations, in which images are segmented horizontally and texts are aligned adaptively. Besides, we introduce a multi-stage cross-modal matching strategy to match the visual and textual representations from three levels and eliminate the modality gap step by step. The outstanding experimental results verify the superiority of our proposed TIPCB method.

\section*{Acknowledgment}
This research is supported in part by the National Natural Science Foundation of China under Grant 61806099, U20B2065; and by the Natural Science Foundation of Jiangsu Province of China under Grant BK20180790.


\end{document}